\documentclass[10pt,twocolumn,letterpaper]{article}

\usepackage{wacv}              

\usepackage{graphicx}
\usepackage{bm} 
\usepackage{amsmath}
\usepackage{amssymb}
\usepackage{booktabs}
\usepackage{makecell}  
\usepackage{multirow}
\usepackage{textcomp} 
\usepackage{comment}
\usepackage[most]{tcolorbox}   
\usepackage{xcolor}    

\usepackage[pagebackref,breaklinks,colorlinks]{hyperref}

\usepackage[capitalize]{cleveref}
\crefname{section}{Sec.}{Secs.}
\Crefname{section}{Section}{Sections}
\Crefname{table}{Table}{Tables}
\crefname{table}{Tab.}{Tabs.}

\def\wacvPaperID{2706} 
\def\confName{WACV}
\def\confYear{2026}

\begin{document}

\title{VAOT: Vessel-Aware Optimal Transport for Retinal Fundus Enhancement}

\author{
Xuanzhao Dong$^{1*}$, Wenhui Zhu$^{1*}$, Yujian Xiong$^{1}$,
Xiwen Chen$^{2}$, Hao Wang$^{2}$, Xin Li$^{1}$, Jiajun Cheng$^{1}$ \\
Zhipeng Wang$^{3}$,  Shao Tang$^{3}$, Oana Dumitrascu$^{4}$, Yalin Wang$^{1}$ \\
$^{1}$Arizona State University, AZ, USA \\
$^{2}$ Clemson University, SC, USA \\
$^{3}$LinkedIn Corporation, CA, USA \\
$^{4}$Mayo Clinic, AZ, USA\\
}
\maketitle
\def\thefootnote{*}\footnotetext{ Contributed equally to this paper.}

\begin{abstract}
Color fundus photography (CFP) is central to diagnosing and monitoring retinal disease, yet its acquisition variability (e.g., illumination changes) often degrades image quality, which motivates robust enhancement methods. Unpaired enhancement pipelines are typically GAN-based, however, they can distort clinically critical vasculature, altering vessel topology and endpoint integrity. Motivated by these structural alterations, we propose Vessel-Aware Optimal Transport (\textbf{VAOT}), a framework that combines an optimal-transport objective with two structure-preserving regularizers: (i) a skeleton-based loss to maintain global vascular connectivity and (ii) an endpoint-aware loss to stabilize local termini. These constraints guide learning in the unpaired setting, reducing noise while preserving vessel structure. Experimental results on synthetic degradation benchmark and downstream evaluations in vessel and lesion segmentation demonstrate the superiority of the proposed methods against several state-of-the art baselines. The code is available at \url{https://github.com/Retinal-Research/VAOT}
\end{abstract}

\section{Introduction}
Retinal color fundus photography (CFP) is pivotal for diagnosing and monitoring ocular diseases~\cite{deeplearning1,deeplearning5,zhu2023beyond,zhu2024nnmobilenet,dumitrascu2024color,zhu2025retinalgpt}, but image quality is often degraded by uncontrolled acquisition factors such as variable ambient illumination, motion or focus errors, and operator-induced artifacts. The resulting degradations (e.g., contrast mismatch, nonuniform illumination, and vessel attenuation) compromise the faithful visualization of lesions and vasculature that is essential for conditions like hypertensive retinopathy. Consequently, developing robust enhancement frameworks for low-quality CFP is critical to improving diagnostic reliability and downstream analysis~\cite{zhu2023otre,zhu2023optimal,dong2024cunsb}.


Unpaired algorithms currently dominate CFP enhancement, as small acquisition or implementation variations introduce subtle but pervasive differences across images, making the collection of paired data impractical. These methods typically cast enhancement as unpaired domain translation between low and high quality distributions. For example, OTTGAN~\cite{wang2022optimal} formulates a Monge optimal transport (OT) objective within an adversarial framework, avoiding the additional computational overhead of auxiliary structures used by CycleGAN~\cite{zhu2017unpaired} and enabling smoother domain shifts. Nonetheless, a central limitation remains when transferring such methods to retinal imaging, which is preserving clinically salient anatomy. In fact, retinal images contain complex structures (e.g., heterogeneous lesions and fine vascular networks) and GAN-based translators, which are prone to mode collapse and unstable training, can distort vessel topology, attenuate subtle lesions or even bring additional artifacts, ultimately undermining their clinical reliability~\cite{zhu2023otre,dong2024cunsb,vasa2024context}. This motivates the incorporation of explicit structure-aware constraints to maintain anatomical fidelity during enhancement.

To address these challenges, we propose Vessel-Aware Optimal Transport (VAOT), a novel vessel-aware retinal fundus image enhancement framework with three complementary components. First, we leverage an OT theory to promote stable, smooth distribution shifts in the unpaired setting. Second, to preserve global vascular topology, we introduce a skeleton-based regularizer that treats the one-pixel-wide vessel skeleton as a structural prior and a proxy for anatomical alignment. Third, to refine local geometry, we add an endpoint-aware regularization term in image space that emphasizes small, clinically relevant structures. Our contributions are threefold:

\begin{itemize}
    \item We introduce VAOT, a vessel-aware retinal fundus enhancement model that improves image quality while preserving diagnostically important vascular structure.
    \item We propose a mixed structure-preserving scheme. Specifically, skeleton consistency to maintain global topology and overall vessel layout. Endpoint-aware constraints to align local termini.
    \item We conduct large-scale experiments on public retinal fundus datasets, showing that VAOT outperform multiply baselines in both denoising performance and preservation of vascular topology.
\end{itemize}

\section{Related Work}
\subsection{Retinal Fundus Image Enhancement}
Early work on retinal fundus image enhancement largely relied on paired supervised or self-supervised formulations. For example, PCE-Net~\cite{10.1007/978-3-031-16434-7_49} leverages Laplacian-pyramid constraints derived from low-quality inputs to guide enhancement, while GFE-Net~\cite{li2023generic} and SCR-Net~\cite{scrnet} inject frequency cues and synthetic data as additional priors. However, the requirement for paired images limits their scalability and practicality in clinical workflows. 

The success of CycleGAN~\cite{zhu2017unpaired} opened a path to unpaired enhancement by introducing a dual-generator architecture with cycle consistency, enabling bidirectional mappings without aligned pairs. However, for CFP enhancement, which is predominantly one-directional (i.e., low to high quality), the auxiliary reverse mapping and extra modules add computational overhead that is often unnecessary. Additionally, building on WGAN-based models~\cite{arjovsky2017wasserstein,gulrajani2017improved}, subsequent approaches~\cite{zhu2023optimal,zhu2023otre} employed optimal-transport (OT) theory to achieve smoother distribution alignment. For instance, context-aware OT~\cite{vasa2024context} uses deep feature spaces (i.e., VGG~\cite{mechrez2018contextual}) to compute contextual losses that approximate Earth Mover’s Distance over representations. Beyond GANs, diffusion-based methods have also shown promise. For example, CUNSB-RFIE~\cite{dong2024cunsb} learns a Schr\"{o}dinger Bridge (SB) between image domains by estimating multiply SB couplings across sub-intervals via adversarial training, yielding smooth, probabilistically consistent transformations. However, this method fall behind GANs, as those high-frequency details (e.g., lesion structure) are gradually smoothed out during iterative training.

Despite these advances, preserving clinically salient vasculature remains a core challenge. Many existing methods prioritize image or feature level perceptual alignment and may under-penalize localized topological errors (e.g., breaks, merges and endpoint shifts) in blood vessels. To our knowledge, VAOT is the first unpaired CFP enhancement framework that explicitly targets vascular preservation with complementary constraints at two scales. Specifically, the skeleton-based regularization targets global morphology, while the endpoint-aware regularization emphasize local structural alignment.

\subsection{Topology Structure Preservation}
Topology preservation has been extensively studied in segmentation, particularly for tubular and curvilinear structures (e.g., retinal vasculature in diabetic retinopathy). For instance, ~\cite{hu2019topology} introduce a differentiable loss based on persistent homology that aligns birth–death pairs in persistence diagrams to maintain correct connectivity and cycles, while SkelCon~\cite{tan2022retinal} integrates skeletal priors into the segmentation network to better preserve vessel morphology. However, despite this progress in segmentation and delineation, applying topology-preserving objectives to retinal fundus image enhancement remains underexplored. VAOT addresses this gap by coupling enhancement with structure-aware constraints that preserve vascular morphology at both global (i.e., overall skeleton ) and local (i.e., endpoint-based window) scales.

\begin{figure}[ht]
  \centering
  \includegraphics[width=1\columnwidth]{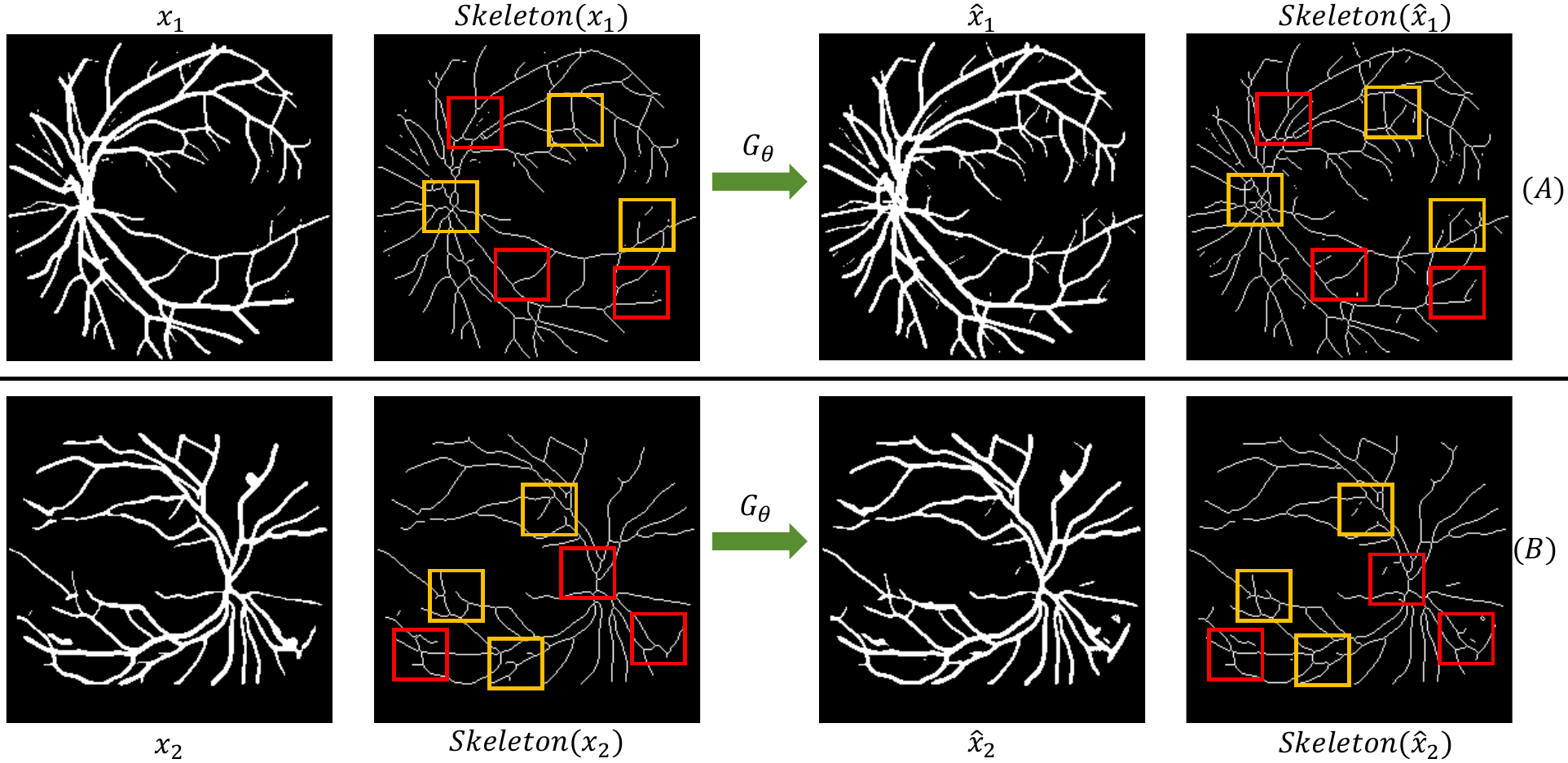}
  \caption{Example \textbf{(A)} and \textbf{(B)} illustration the segmentation mask and corresponding skeleton map (e.g., \textit{Skeleton}($\mathbf{x}_1$)) for input low quality images (e.g., $\mathbf{x}_1$) and its enhanced counterparts (e.g., $\hat{\mathbf{x}}_1$). The skeleton structure shifts and endpoint modifications are highlight in yellow and red box, respectively. Here, the freezed generator $G_\theta$ comes from~\cite{zhu2023optimal} considering it's consistent great performance shown in~\cite{zhu2025eyebench}, and segmentation network comes from~\cite{zhou2021study}. Algorithms outlined in~\cite{van2014scikit} are used to extract skeleton based on segmentation map. See Sec.~\ref{sec:preliminary} for more detailed analysis.
  }
  \label{fig:motivation}
\end{figure}

\section{Preliminary and Analysis}\label{sec:preliminary}
Let $\mathbf{X}\sim \mathbb{P}_{\mathbf{X}}d\mu$ and $\mathbf{Y}\sim \mathbb{P}_{\mathbf{Y}}d\nu$ denote low-quality and high-quality fundus images, respectively, where $\mu, \nu$ represent the corresponding measure. We consider two disjoint, unpaired datasets $\mathcal{D}_{\mathbf{X}}=\{\mathbf{x}_i\}_{i=1}^{N_X}$ and $\mathcal{D}_{\mathbf{Y}}=\{\mathbf{y}_j\}_{j=1}^{N_Y}$ drawn independently from $\mathbb{P}_{\mathbf{X}}$ and $\mathbb{P}_{\mathbf{Y}}$, with no one-to-one correspondences. VAOT, which belong to unpaired enhancement framework, is cast as image-to-image (I2I) translation task. Specifically, it learn a mapping $G_{\theta}\!:\!\mathbf{X}\!\to\!\mathbf{Y}$ such that
\[
\hat{\mathbf{x}}_i \;=\; G_{\theta}(\mathbf{x}_i)
\]
matches the target distribution $\mathbb{P}_{\mathbf{Y}}$ while preserving the vascular morphology present in $\mathbf{x}_i$. Here, $G_{\theta}$ is modeled as a generator parameterized by $\theta$.
Existing retinal enhancement frameworks can fail to preserve vascular morphology. In particular, vascular topology may be altered, manifesting as global skeleton changes (e.g., formation or loss of loops) and local endpoint deviations (e.g., spurious branches or vessel breaks). As illustrated in Fig.~\ref{fig:motivation} (A and B), when applied to low-quality inputs, the generator $G_\theta$ can introduce artifacts such as false-positive branches (e.g., highlighted in yellow box). The red box marks regions where endpoint conservation is violated (e.g., newly introduced discontinuities). Although visually subtle (i.e., the overall segmentation may appear well preserved), these perturbations substantially degrade structural fidelity and can impair downstream clinical tasks.
Directly porting topology-preserving methods from segmentation to enhancement is nontrivial due to mismatched supervision (i.e., explicit mask labels vs. implicit image intensities). As a result, many retinal enhancement methods rely on global regularization (e.g., image- or feature-level perceptual losses) that emphasize overall similarity but under-penalize small structural defects (e.g., terminal-branch breaks or short spurious vessels). The issue is exacerbated when regularizing with segmentation masks, since noisy inputs and imperfect labels can compromise network integrity, making it difficult to distinguish genuine vascular structure from false positives. As a result, hard alignment to such masks may even hinder denoising. This raises a central design question for vessel-aware enhancement:

\textcolor{red}{\textit{Given limited and potentially noisy structural cues (e.g., segmentation masks), how can we enforce both global and local vascular morphology preservation during enhancement?}}

To address these issues, we propose VAOT, which comprises three complementary components detailed in the next section.

\section{Method}
VAOT comprises three modules: (i) an optimal-transport–based enhancement backbone that induces smooth distribution shifts in the unpaired setting (Sec.~\ref{subsec:ot-enhancement}); (ii) Skeleton-Guided global morphology Alignment (SGA), a regularizer designed to preserve global vascular topology using skeleton pseudo-labels (Sec.~\ref{subsec:sga}); and (iii) Endpoint-based vascular integrity Preservation (EVP), which refines local structure around vessel terminals in image space (Sec.~\ref{subsec:endpoint}).

\subsection{Image Enhancement Guided by Optimal Transport}\label{subsec:ot-enhancement}
As discussed in Sec.~\ref{sec:preliminary}, the retinal image enhancement problem can be cast as a \textit{Monge’s} optimal transport problem, denoted as:
\begin{equation} \label{eq:ot}
    \begin{split}
        f^*  &= \inf_f \int_\mathbf{X} C(\mathbf{x}, f (\mathbf{x}) )d \mu\\
        &\text{subject to} \ f_{\#}\mu = \nu
    \end{split}
\end{equation}
Here, $f$ denote a measurable transport map and $C(\cdot, \cdot)$ is a pointwise cost function (e.g., the squared L2 loss). The push forward constraint $f_{\#}\mu = \nu$ ensure that for any measurable set $B \subseteq \mathbf{Y}$, the probability mass of $f^{-1}(B) \subseteq\mathbf{X}$ is preserved.

Following prior works~\cite{zhu2023otre,zhu2023optimal,wang2022optimal}, we model Eq.~\ref{eq:ot} with GAN-style adversarial learning framework, denoted as:
\begin{equation}\label{eq:ot-objective}
    \begin{split}
        \max_{G_\theta} \min_{D_\beta} \quad&\lambda_1\mathbb{E}_{\mathbf{x}} [ C( \mathbf{x}, G_\theta (\mathbf{x}) )]  + \lambda_2\mathbb{E}_{\mathbf{y}} [ C( \mathbf{y}, G_\theta (\mathbf{y}) )] + \\
        & \mathbf{W}_1(\mathbb{P}_{G_\theta(\mathbf{X})}, \mathbb{P}_\mathbf{Y})\\
    \end{split}
\end{equation}
Specifically, the generator $G_\theta$ aims to approximate optimal transport map $f^*$, while the discriminator $D_\beta$ serves as a 1-Lipschitz critic estimating the Wasserstein-1 distance between the generated distribution $\mathbb{P}_{G_\theta(\mathbf{X})}$ and the target distribution $\mathbb{P}_\mathbf{Y}$. The identity regularization term $C( \mathbf{y}, G_\theta (\mathbf{y}))$ discourages unnecessary changes to already clean images. $\lambda_1$ and $\lambda_2$ works as balancing parameters. We instantiate cost function $C$ using the Local Quasi-Convex Multi-Scale Structural Similarity Index Measure~\cite{wang2003multiscale,brunet2011mathematical}, and we enforce the Lipschitz condition on $D_\beta$ with the gradient penalty method~\cite{gulrajani2017improved}.

However, strong distributional alignment does not by itself ensure preservation of vascular morphology (see example (B) in Fig.~\ref{fig:motivation}), even under objectives that emphasize perceptual similarity (e.g., SSIM). Additional mechanisms are needed to encourage both global topology and local structural alignment without compromising denoising performance.
\begin{figure}[ht]
  \centering
  \includegraphics[width=1\columnwidth]{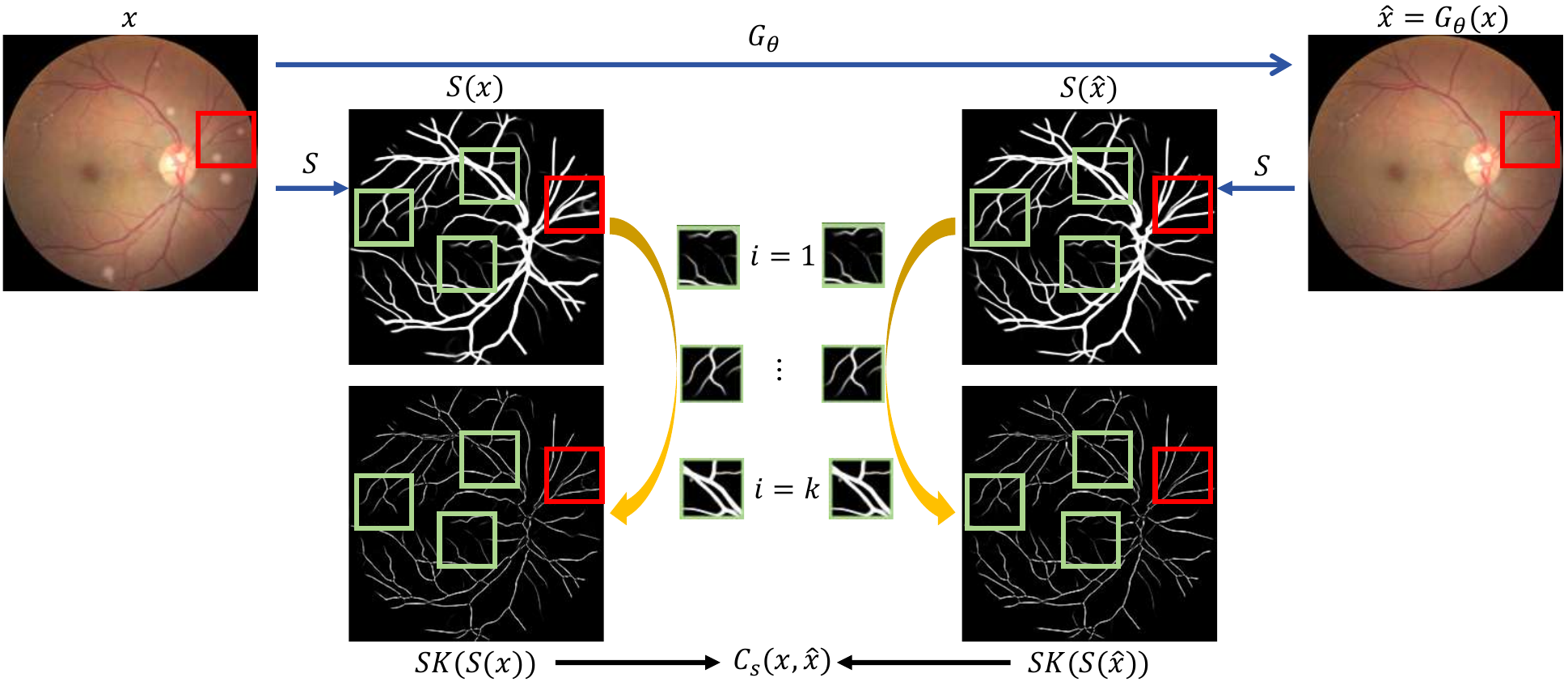}
  \caption{Illustration of the SGA modules. Specifically, given low quality image $\mathbf{x}$ and its enhanced counterpart $\hat{\mathbf{x}}=G_\theta(\mathbf{x})$, we use their soft segmentation maps (i.e., sigmoid outputs) and the corresponding soft skeletons to regularize global shape and connectivity. The red box highlights regions where noise affects the results, and the green box illustrates the order of skeletonization sequence. As the step index $i$ increases, centerlines of progressively thicker vessels are extracted. See Sec.~\ref{subsec:sga} for details.
  }
  \label{fig:SGA}
\end{figure}

\subsection{Skeleton-Guided Global Morphology Alignment}\label{subsec:sga}
Give a low quality image $\mathbf{x}$ and its enhanced counterpart $\hat{\mathbf{x}}$, we can get the corresponding segmentation likelihoods $S(\mathbf{x})$ and $S(\hat{\mathbf{x}})$ (e.g., after sigmoid operation) from a pretrained vessel segmenter $S$. However, enforcing topology with these masks is risky, as the noise and artifacts in $\mathbf{x}$ often degrade $S(\mathbf{x})$, so forcing $S(\hat{\mathbf{x}})$ to match $S(\mathbf{x})$ can anchor the generator to noisy labels and conflict with denoising. As shown in Fig.~\ref{fig:SGA}, the red box highlight regions where noise affects the results, directly enforcing alignment will encourage $G$ preserve spurious loops.

In contrast, the morphological skeleton $SK(\cdot)$ (e.g., $SK(\mathbf{x})$), which represents vessels as one-pixel-wide centerlines, provides a compact alternative that preserves vascular connectivity and branching while discarding uncertain boundaries. This representation is inherently more robust to blur, low contrast, and label jitter that commonly corrupt full masks and exist in noisy images. Even when masks are imperfect, their induced skeletons offer a reliable topological prior for morphology preservation during enhancement (see Fig.~\ref{fig:SGA}). However, the differentiability present to be a practical caveat, considering the hard skeletonization is non-differentiable.

Soft-Skeletonization~\cite{shit2021cldice} provide a solution, which approximate morphological skeleton on a probability mask using min and max pooling. Specifically, given the current map $R_i$ (e.g., $R_0 := S(\mathbf{x})$) and $E_{i} = minpool(R_i)$ as the soft erosion process in step $i$. The algorithm will compute the opening $O_i$ such that $O_i := maxpool(E_i)$. Then, the i-step soft skeleton layer $SK_i := ReLU (R_i - O_i)$ captures the centerline remnants at that radius. Intuitively, as shown in Fig.~\ref{fig:SGA}, each erosion step shrinks the local vessel radius, and the following opening regrows smooth interior regions. As a result, only vessel regions that are thick enough at depth $i$ survive, with the remaining structure (i.e., $SK_i$) constitutes the per-depth skeleton layer (i.e., the centers of maximal inscribed disks). As the process continues, layers corresponding to progressively thicker vessels are extracted. (shown in green box in Fig.~\ref{fig:SGA}). Naturally, the final skeleton $SK$ is the union of all per-depth layers, denoted as $SK(S(\mathbf{x})):=\bigcup_{i=0}^{k}SK_i$. Here, $k$ represent the hyperparameter of total iteration steps.

Because we apply soft segmentation $S(\cdot)$ and compose only differentiable pooling operations and ReLU activation, the resulting soft skeleton $SK(\cdot)$ is naturally differentiable. We then define the skeleton-guided global topology preservation regularization (SGA) as:
\begin{equation}\label{eq:cldice}
\begin{aligned}
C_s
& := 1-\frac{2\times T_p(SK(S(\hat{\mathbf{x}})), S(\mathbf{x})) \times T_s(SK(S(\mathbf{x})), S(\hat{\mathbf{x}}) )}{T_p(SK(S(\hat{\mathbf{x}})), S(\mathbf{x})) + T_s(SK(S(\mathbf{x})), S(\hat{\mathbf{x}}) )} \\
&\text{where }\left\{
\begin{aligned}
T_p &= \frac{\sum\!\big(SK(S(\hat{\mathbf{x}})) \odot S(\mathbf{x})\big)+\epsilon}
             {\sum\!SK(S(\hat{\mathbf{x}})) + \epsilon} \\[4pt]
T_s &= \frac{\sum\!\big(SK(S(\mathbf{x})) \odot S(\hat{\mathbf{x}})\big)+\epsilon}
             {\sum\!SK(S(\mathbf{x})) + \epsilon}
\end{aligned}
\right.
\end{aligned}
\end{equation}
Here $T_p$ asks whether predicted centerlines lie inside high-probability input vessel regions, $T_s$ checks whether the prediction covers the input's centerlines, $\epsilon$ ensures numerical stability and $\odot$ works as element-wise production. Minimizing this loss penalizes spurious centerlines and gaps in prediction, which promotes preservation of overall vessel shape and connectivity~\cite{shit2021cldice}. 

\begin{figure}[ht]
  \centering
  \includegraphics[width=\columnwidth]{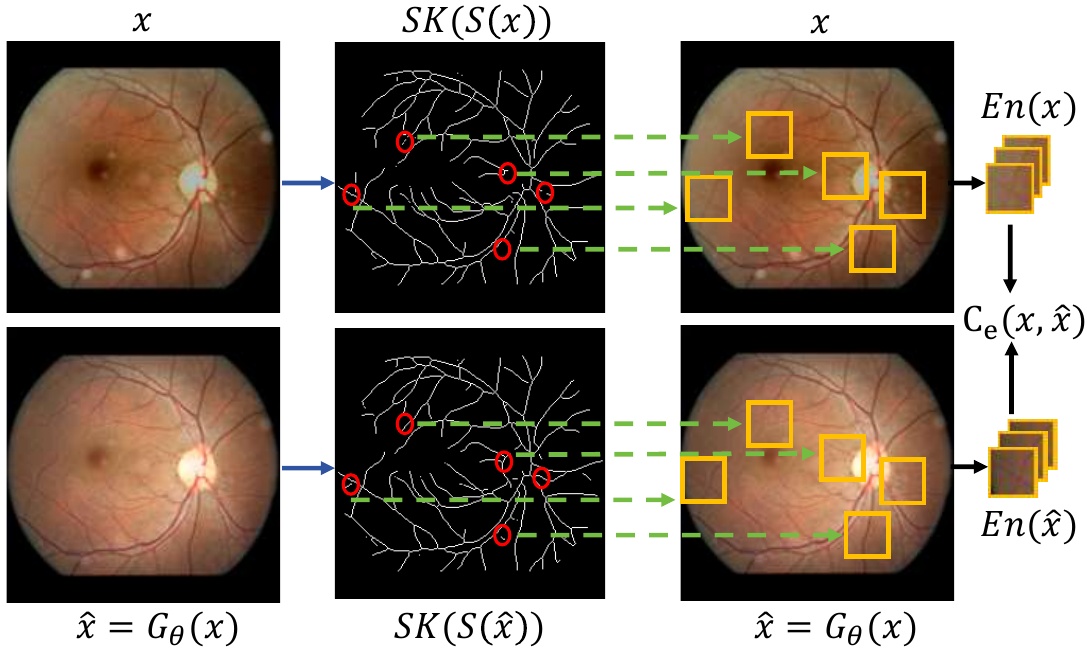}
  \caption{
  Illustration of the EVP module. Red circles mark vessel endpoints derived from the vessel skeleton, yellow boxes show the endpoint-centered windows in image space, and green dashed arrows indicate their correspondences. The local morphology regularization $C_e$ is computed over these batches of small, endpoint-centered windows. We denote this extraction-and-windowing operator by $En(\cdot)$. For simplicity, we overload the skeletonization notation $SK (\cdot)$ to denote a hard, binary skeleton. See Sec.~\ref{subsec:endpoint} for details.
  }
  \label{fig:EVP}
\end{figure}

\subsection{Local Vascular Integrity via Endpoint Priors}\label{subsec:endpoint}
Preserving vascular structure with a global, skeleton-based regularizer alone is not sufficient. Local topological changes (e.g., endpoint creation or loss) require additional attention. Given a low-quality image $\mathbf{x}$, we obtain its skeleton $SK(S(\mathbf{x}))$. For endpoint detection, we intentionally use a hard, binary skeleton and we overload $SK(\cdot)$ and $S(\cdot)$ to denote binarized operators here for simplicity.

Based on binary skeletonization, endpoints can be localized with a simple neighborhood count through all-ones $3\times3$ convolution kernel. Specifically, for each skeleton pixel, we count the number of skeleton pixels in its 8-connected neighborhood, and the endpoints are those pixels whose result is 1 (e.g., endpoints in marginal vessel branches or breaks) or 0 (e.g., the isolated points) besides the pixel itself. As illustrated in Fig.~\ref{fig:EVP}, red circles mark endpoint differences that arise after enhancement. In practice, such endpoint shifts typically occur in three regions: (i) peripheral, very thin branches, (ii) areas affected by image noise or artifacts, and (iii) neighborhoods of major anatomical structures (e.g., the optic disc), where brightness and texture can perturb segmentation and skeletonization. However, directly evaluating endpoints is challenging because they are represented as single-pixel, binary locations in skeleton. Moreover, unlike segmentation tasks that typically provide dense ground-truth masks, endpoint shifts observed in the enhancement setting may legitimately arise from artifact removal (as in  Fig.~\ref{fig:EVP}) and should sometimes be treated as correct behavior rather than errors. Therefore, endpoint-based local structure preservation requires careful design.

Our approach, named EVP, is intuitive: rather than comparing endpoints directly, we regularize local structure within small windows centered at detected endpoints. In this setup, endpoints act only as anchors to localize regions that warrant additional attention. As shown in Fig.~\ref{fig:EVP}, when an endpoint is selected (red circle), we extract an $l\times l$ square window centered at that location in both the low-quality image $\mathbf{x}$ and its enhanced counterpart $\hat{\mathbf{x}}$. Let $M$ denote the set of those endpoint with $m := |M|$, and define $En_j(\cdot)$ represent the operation that crops the j-th endpoint-centered window. Then our endpoint-aware local regularizer defined as:
\begin{equation}\label{eq:evp}
    C_e(\mathbf{x}, \hat{\mathbf{x}}) :=\frac{1}{m}\sum_{j=1}^{m}C (En_j(\mathbf{x}), En_j(\mathbf{\hat{x}}))
\end{equation}
where $C$ is an SSIM-based loss, designed to preserve local perceptual structure, and Eq.~\ref{eq:evp} can be easily implemented to batch of images. Our design bring three advantages. First, this formulation avoids brittle, one-pixel endpoint matching. Second, regardless of whether an endpoint appears to shift due to artifact or denoising, the model is always encouraged to maintain coherent local morphology around those locations, which susceptible to structure changes. Finally, because the loss is applied to image patches, it remains differentiable with respect to the generated image. We use the algorithms outlined in~\cite{zhou2025glcp} to organize the endpoints selection. 
\begin{figure}[h]
  \centering
  \includegraphics[width=0.95\columnwidth]{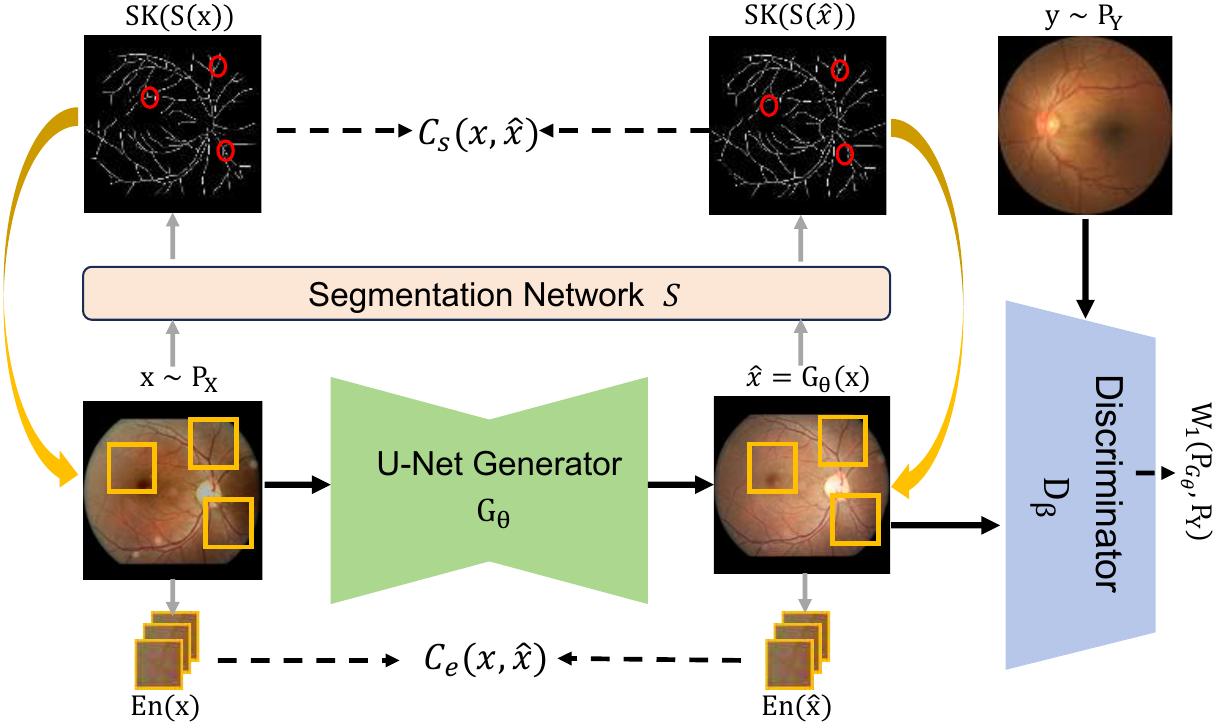}
  \caption{
  Illustration of the VAOT pipeline. The operator $SK(\cdot)$ denotes skeletonization, which is soft in Phase 1 and hard in Phase 2. Red circles mark detected endpoints and yellow arrows indicate the corresponding endpoint-centered local windows in image space. See Sec.~\ref{subsec:final-target} for details.
  }
  \label{fig:structure}
\end{figure}

\subsection{Main learning Objective}~\label{subsec:final-target}
Training VAOT has two phases. As shown in Fig.~\ref{fig:structure}, Phase 1 trains for $T_1$ epochs with the OT-based adversarial objective (i.e., Eq.~\ref{eq:ot-objective}) to learn distribution transport while avoiding unnecessary changes on clean images. Phase 2 adds the SGA and EVP regularizers to preserve global morphology and local structure, and continues training for $T_2$ epochs with:
\begin{equation}\label{eq:ot-objective-final}
    \begin{split}
        \max_{G_\theta} \min_{D_\beta} \quad &\lambda_1\mathbb{E}_{\mathbf{x}} [ C( \mathbf{x}, G_\theta (\mathbf{x}) )]  + \lambda_s\mathbb{E}_{\mathbf{x}} [ C_s( \mathbf{x}, G_\theta (\mathbf{x}) )] \\
        & + \lambda_e \mathbb{E}_{\mathbf{x}} [ C_e( \mathbf{x}, G_\theta (\mathbf{x}) )] + \mathbf{W}_1(\mathbb{P}_{G_\theta(\mathbf{X})}, \mathbb{P}_\mathbf{Y})
    \end{split}
\end{equation}
Here, $\lambda_s$ and $\lambda_e$ balance the topology terms. $C$ denotes the base cost function, $C_s$ is the skeleton-guided global topology term (i.e., Eq.~\ref{eq:cldice}), and $C_e$ is the endpoint-aware local regularizer (i.e., Eq.~\ref{eq:evp}).

\begin{table*}[!t]
\centering
\caption{Performance comparison of denoising evaluation in Full-Reference quality assessment experiments. The best performance in each column is highlighted in bold, with the second-best underlined.}
\tiny
\resizebox{1\textwidth}{!}{%
\begin{tabular}{cccccccc}
\toprule
\multirow{2}{*}{} & \multirow{2}{*}{\textbf{Method}} & \multicolumn{2}{c}{\textbf{EyeQ}} & \multicolumn{2}{c}{\textbf{IDRID}} & \multicolumn{2}{c}{\textbf{DRIVE}} \\ \cmidrule(l){3-8} 
                                          &                                  & \textbf{SSIM} $\uparrow$   & \textbf{PSNR} $\uparrow$   & \textbf{SSIM} $\uparrow$   & \textbf{PSNR} $\uparrow$   & \textbf{SSIM} $\uparrow$   & \textbf{PSNR} $\uparrow$   \\ \midrule
\multirow{5}{*}{\textit{Paired Methods}} & SCR-Net~\cite{li2022structure}   & \textbf{0.9606} & 29.698 & 0.6425 & 18.920 & \textbf{0.6824} & 23.280 \\      
                                         & Cofe-Net~\cite{shen2020modeling} & 0.9408           & 24.907           & 0.7397            & 20.058            & 0.6671            & 21.774            \\
                                         & PCE-Net~\cite{10.1007/978-3-031-16434-7_49} & 0.9487           & \textbf{29.895}           & \underline{0.7764}            & \underline{23.201}           & 0.6704            & 24.041           \\
                                         & GFE-Net~\cite{li2023generic}     & \underline{0.9554}           & \underline{29.719}           & \textbf{0.7935}            & \textbf{25.012}           & \underline{0.6793}            & \underline{23.786} \\
                                         &RFormer~\cite{deng2022rformer}
                                          & 0.9260   & 27.163   & 0.5963   & 18.433   & 0.6311   & 22.172\\
                                          \midrule
\multirow{7}{*}{\textit{Unpaired Methods}} 
                                         & CycleGAN~\cite{zhu2017unpaired}         & 0.9313         &\underline{25.076}           & 0.7668            & \underline{22.511}           & \underline{0.6681}            & \textbf{22.686}  \\
                                         & WGAN~\cite{gulrajani2017improved} & 0.9266          & 24.793           & 0.7316           & 21.325           & 0.6431            & 20.408 \\
                                         & OTTGAN~\cite{wang2022optimal}    & 0.9275           & 24.065           & 0.7509           & 22.131           & 0.6635            & 21.938 \\
                                         & OTEGAN~\cite{zhu2023optimal}     & \underline{0.9392}           & 24.812         & 0.7624           & 22.272           & 0.6642            & 22.183 \\
                                         & Context-aware OT~\cite{vasa2024context} & 0.9144           & 24.088           & 0.7338            & 21.790            & 0.6407            & 21.389 \\
                                         & CUNSB-RFIE~\cite{dong2024cunsb}  & 0.9121           & 24.242           & \underline{0.7651}           & 22.448           & 0.6659            & \underline{22.510} \\
                                         \midrule
                                         & VAOT(ours) & \textbf{0.9433} & \textbf{25.253} & \textbf{0.7674} & \textbf{22.793} & \textbf{0.6701} & 22.287\\
\bottomrule
\end{tabular}%
}
\label{tab:mainexp}
\end{table*}

\begin{table*}[!t]
    \centering
    \caption{
    Performance comparison of vessel and lesion (EX and HE) segmentation in main enhancement task. In both paired and unpaired algorithms, the best performance in each column is highlighted in bold, with the second-best underlined.}
    \tiny
    \resizebox{1\textwidth}{!}{%
    \begin{tabular}{lcccc|ccc|ccc}
        \toprule
         \multirow{2}[3]{*}{Method} & \multicolumn{4}{c}{Vessel Segmentation} & \multicolumn{3}{c}{EX} & \multicolumn{3}{c}{HE} \\ 
         \cmidrule(lr){2-5}  \cmidrule(lr){6-8}  \cmidrule(lr){9-11}
          
         & AUC $\uparrow$ & PR $\uparrow$ & F1 Score $\uparrow$ & SP $\uparrow$ & AUC  & PR & F1 Score  & AUC & PR & F1 Score \\ \midrule

        SCR-Net~\cite{li2022structure}   & \textbf{0.9227} & \textbf{0.7783} & \textbf{0.7000} & 0.9787 & \textbf{0.9683} & \textbf{0.6041} & \textbf{0.5556} & 0.9377 & 0.3213 & 0.3725\\ 
        cofe-Net~\cite{shen2020modeling} & \underline{0.9188} & \underline{0.7698} & 0.6895 & 0.9801 & 0.9623 & 0.5620 & 0.5349 & 0.9302 & 0.3152 & 0.3281\\
        PCE-Net~\cite{10.1007/978-3-031-16434-7_49} & 0.9146 & 0.7616 & 0.6790 & \textbf{0.9814} & \underline{0.9667} & \underline{0.5876} & 0.5066 & \underline{0.9545} & \underline{0.3639} & \underline{0.3736}\\
        GFE-Net~\cite{li2023generic} & 0.9175 & 0.7669 & 0.6832 & \textbf{0.9814} & 0.9560 & 0.5548 & \underline{0.5380} & \textbf{0.9577} & \textbf{0.4113} & \textbf{0.3751}\\
        RFormer~\cite{deng2022rformer} & 0.8990 & 0.7239 & 0.6374 & \underline{0.9806} & 0.9626 & 0.5593 & 0.4692 & 0.9207 & 0.2677 & 0.3136 \\
    \midrule
        CycleGAN~\cite{zhu2017unpaired} & 0.9015 & 0.7278 & 0.6462 & 0.9801  & 0.9447& 0.4843 & 0.4790 & 0.8970 & 0.1624 & 0.2227\\
        WGAN~\cite{gulrajani2017improved} & 0.9081 & 0.7494 & 0.6768 & 0.9764  & 0.9522 & 0.4942 & 0.4859 & 0.8990 & 0.1847 & 0.2476\\
        OTTGAN~\cite{wang2022optimal} & 0.9034 & 0.7400 & 0.6609 & \textbf{0.9812}  & 0.9492 & 0.4214 & 0.4365 & 0.8179 & 0.1448 & 0.2233 \\
        OTEGAN~\cite{zhu2023optimal} & 0.9156 & \underline{0.7678} & \underline{0.6919} & 0.9797 & 0.9562 & \underline{0.5191} & \underline{0.4868} & \textbf{0.9359} & \textbf{0.2800}& \underline{0.3165} \\
        Context-aware OT~\cite{vasa2024context} & 0.8871 & 0.7077 & 0.6377 & 0.9791 & 0.9305 & 0.3318 & 0.3707 & 0.8091 & 0.0646 & 0.1184 \\
        CUNSB-RFIE~\cite{dong2024cunsb}  & \underline{0.9163} & 0.7626 & 0.6872 & 0.9784 & \underline{0.9572}& \textbf{0.5381} & \textbf{0.4883} & 0.8488 & 0.1489 & 0.1893 \\
        \midrule
        VAOT(Ours) & \textbf{0.9184} & \textbf{0.7742} & \textbf{0.6966} & \underline{0.9811} & \textbf{0.9608} & 0.5043 & 0.4595 & \underline{0.9256} & \underline{0.2680} & \textbf{0.3280}\\
        \bottomrule
    \end{tabular}}
    \label{tab:seg}
\end{table*}

\section{Experiments}
\subsection{Experiment Setup}
\noindent \textbf{Tasks, Baselines and Metrics.} We evaluate VAOT from four aspects: (i) Main enhancement to assess denoising performance; (ii) Cross-dataset evaluation to assess generalization ability; (iii) Downstream vessel segmentation to assess preservation of vascular morphology after enhancement; and (iv) Downstream Exudates (EX) and Hemorrhages (HE) lesion segmentation  to assess retention of fine, high-frequency details post-enhancement.

We compare our method with the baselines in~\cite{zhu2025eyebench}, including \textit{paired algorithms}: SCR-Net~\cite{li2022structure}, cofe-Net~\cite{shen2020modeling}, PCE-Net~\cite{10.1007/978-3-031-16434-7_49}, GFE-Net~\cite{li2023generic}, RFormer~\cite{deng2022rformer}; \textit{unpaired algorithms}:  CycleGAN~\cite{zhu2017unpaired}, OTT-GAN~\cite{wang2022optimal}, OTRE-GAN~\cite{zhu2023otre}, WGAN~\cite{arjovsky2017wasserstein}, Context-aware OT~\cite{vasa2024context} and CUNSB-RFIE~\cite{dong2024cunsb}. In the main enhancement task, we utilize Structural Similarity Index Measure (SSIM) and Peak Signal-to-Noise Ratio (PSNR) as evaluation metrics. For the downstream vessel segmentation task, we assess performance using Area under the ROC Curve (AUC), Area under the Precision-Recall Curve (PR), F1-score and Specificity (SP). In lesion segmentation task, we use AUC, PR and F1-score as metrics. For all experiments, we follow the implementation proposed in~\cite{zhu2025eyebench} for fair comparision.

\noindent \textbf{Dataset.} For the main enhancement task, we use the EyeQ dataset. Following the degradation pipeline in~\cite{shen2020modeling}, we synthesize degraded inputs by combining spot artifacts, illumination perturbations, and blur. The dataset contains 5000 training images, 600 validation images, and 6217 test images. For paired methods, we organize the training set as aligned low–high image pairs. For unpaired methods, the low-quality images are generated from an additional set of 5000 retinal images disjoint from the high-quality set. For downstream evaluation, we use DRIVE~\cite{1282003} for vessel segmentation and IDRiD~\cite{idrid} for lesion segmentation. 

\noindent \textbf{VAOT setting.} Our backbone follow the design outlined in~\cite{zhu2023otre,zhu2023optimal}, and the pretrained vessel segmentation network used to produce soft probability maps follows~\cite{zhou2021study}. Training uses two phases of $T_1=T_2 = 100$ epochs. At the start of each phase, the learning rate is reset to $1 \times 10^{-4}$ and scheduled with cosine annealing. For soft-skeletonization we use $k=20$ iterations and a numerical stabilizer $\epsilon=1\times10^{-3}$. Additionally, we set the local window size to be 64 for our EVP module. For loss weight, we set $\lambda_1 = 30, \lambda_2 = 15, \lambda_s = 1, \lambda_e = 35$.

\begin{figure*}[!t]
  \centering
  \includegraphics[width=0.85\textwidth]{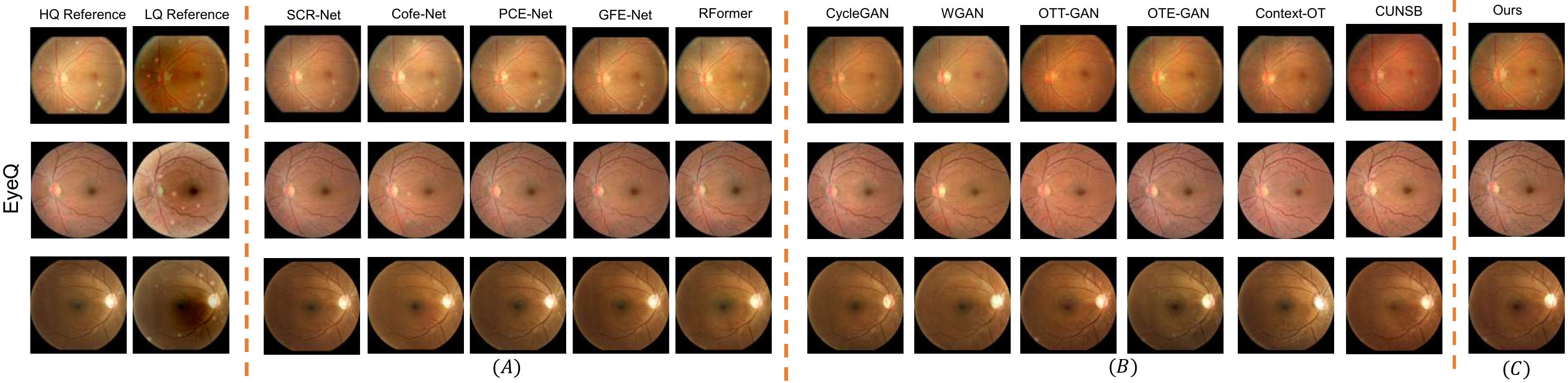}  
  \caption{Illustration of main enhancement task over EyeQ. The first two columns represent low-high quality image pairs. Column \textbf{(A)} illustrates results from paired algorithms, column \textbf{(B)} shows results from unpaired algorithms, and column \textbf{(C)} shows our results (i.e., VAOT). See Sec.~\ref{subsec:result} for details.} 
  \label{fig:eyeq}
\end{figure*}

\begin{figure*}[!t]
  \centering
  \includegraphics[width=0.85\textwidth]{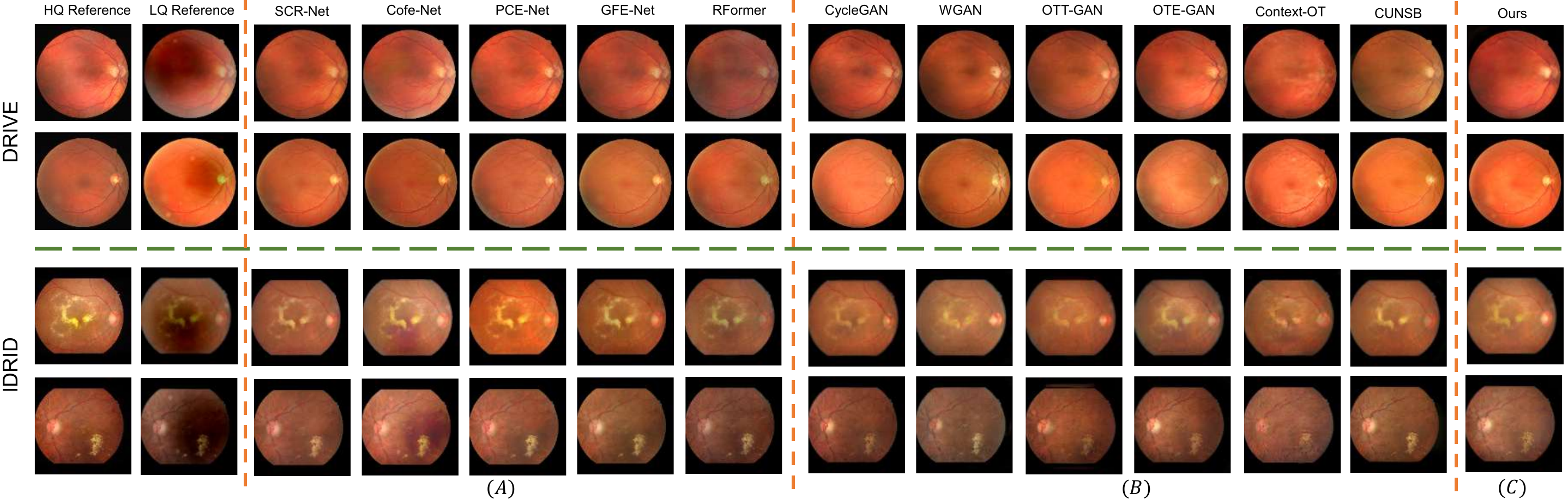}  
  \caption{Illustration of generalization evaulation over IDRID and DRIVE. The enhanced images are generated based on the best weight over EyeQ. The first two columns represent low-high quality image pairs. Column \textbf{(A)} illustrates results from paired algorithms, column \textbf{(B)} shows results from unpaired algorithms, and column \textbf{(C)} shows our results (i.e., VAOT). See Sec.~\ref{subsec:result} for details.} 
  \label{fig:general}
\end{figure*}


\begin{figure}[ht]
  \centering
  \includegraphics[width=0.9\columnwidth]{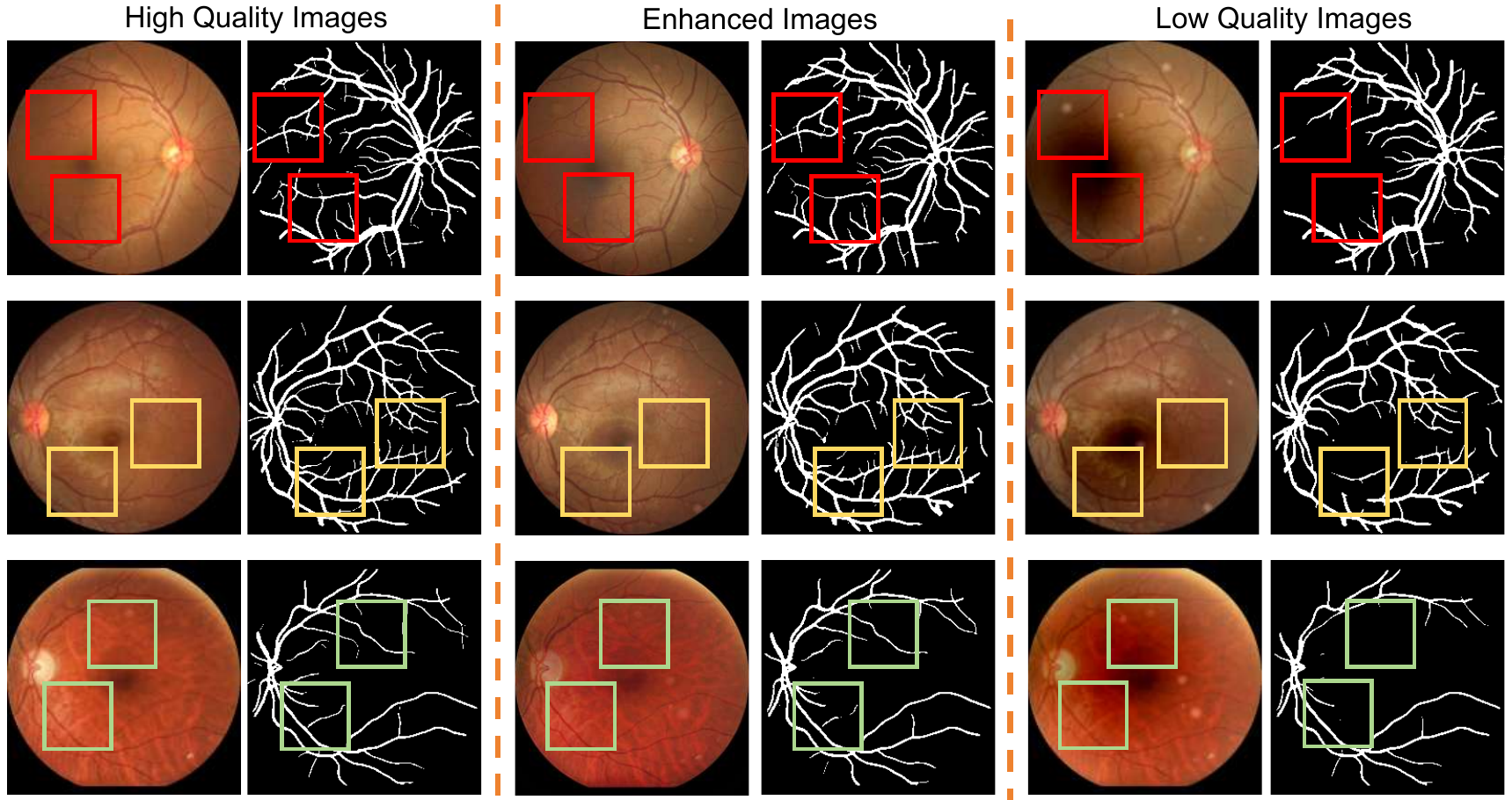}  
  \caption{Illustration of high-quality, low-quality, and enhanced fundus images on EyeQ test set. Corresponding vessel-segmentation maps are shown alongside, generated using the model of~\cite{zhou2021study}. Red, yellow, and green boxes highlight salient differences in vessel topology. See Sec.~\ref{subsec:result} for details.} 
  \label{fig:eyeq-seg}
\end{figure}

\subsection{Experiment Results}\label{subsec:result}
\noindent \textbf{Main Enhancement Tasks.}
Among all unpaired algorithms, VAOT demonstrates superior performance on the main enhancement task and in cross-dataset generalization on IDRID and DRIVE. We train on EyeQ and apply the resulting model without adaptation to enhance low-quality images synthesized with the same degradation pipeline on IDRID and DRIVE. As shown in Tab.~\ref{tab:mainexp}, VAOT achieve the best SSIM (0.9433) and PSNR (25.253) over EyeQ, the highest SSIM and PSNR on IDRID and the highest SSIM on DRIVE. Representative visual comparisons are shown in Fig.~\ref{fig:eyeq} (EyeQ) and Fig.~\ref{fig:general} (IDRiD and DRIVE), which qualitatively corroborate the quantitative gains.

To evaluate whether our SGA and EVP designs help vascular structure preservation, we further analysis the vessel segmentation results from low-quality inputs, high-quality references and VAOT outputs. As shown in Fig.~\ref{fig:eyeq-seg}, the low-quality images exhibit broken branches and artifact-induced distortion (see red, yellow and green boxes), whereas VAOT restores branch continuity and global vascular shape to closely math the high-quality references. Mechanistically, the SGA term acts as a light global constraint that stabilizes connectivity and coarse geometry based on skeleton, while the EVP term applies an endpoint-based local window in image space to further refine fine structures near terminals. Importantly, since VAOT avoids hard morphology matching to degraded inputs, this topology guidance does not undermine denoising performance (see Tab.~\ref{tab:mainexp}).

Although paired methods can yield higher absolute scores in Tab.~\ref{tab:mainexp}, this does not diminish our contribution. In real clinical workflows, collecting perfectly aligned low/high-quality pairs is rarely feasible, making unpaired learning the more realistic setting. We therefore report paired baselines for context and completeness, while emphasizing the practical value of our unpaired approach.

\begin{figure*}[!t]
  \centering
  \includegraphics[width=0.8\textwidth]{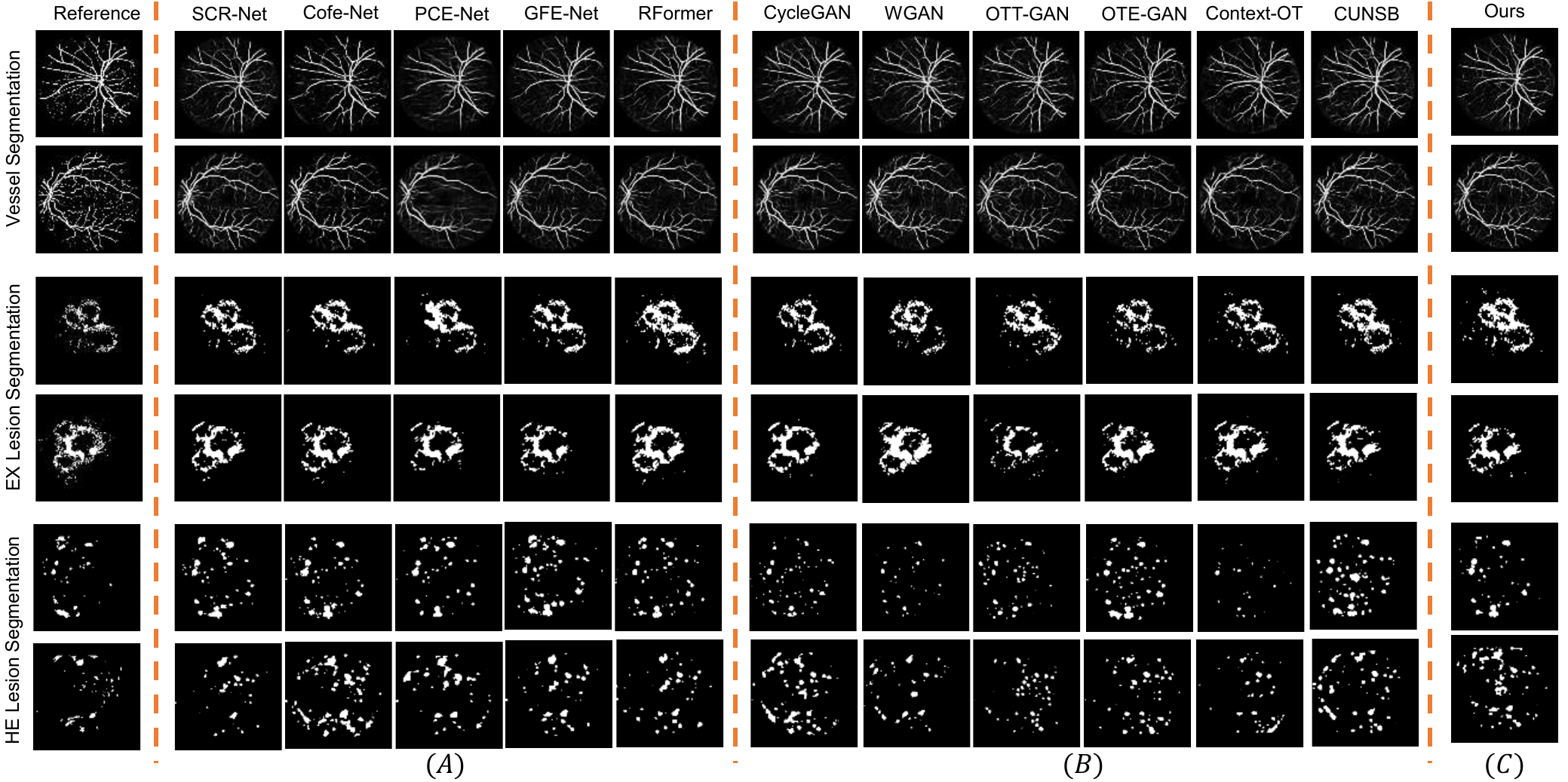}  
  \caption{Illustration of vessel and lesion (i.e., EX and HE) segmentation results over DRIVE and IDRID, respectively. The first columns represent ground truth mask. Column \textbf{(A)} illustrates results from paired algorithms, column \textbf{(B)} shows results from unpaired algorithms, and column \textbf{(C)} shows our results (i.e., VAOT). See Sec.~\ref{subsec:result} for details.} 
  \label{fig:main-segmentation}
\end{figure*}

\noindent \textbf{Vessel and Lesion Segmentation.} Downstream vessel segmentation results further demonstrate effective preservation of vascular morphology. As reported in Tab.~\ref{tab:seg}, VAOT achieves the best AUC, PR, and F1 scores, and the second-best SP score among the seven unpaired methods. Qualitative examples in Fig.~\ref{fig:main-segmentation} corroborate these findings: several unpaired baselines (e.g., CycleGAN, OTT-GAN) introduce artifacts that lead to false-positive vessel predictions or missed peripheral loops, whereas \textsc{VAOT} maintains coherent branching and continuity.

For lesion segmentation (i.e., EX and HE lesion segmentation), \textsc{VAOT}, despite emphasizing vascular structure, still retains strong performance on these higher-frequency details. As shown in Tab.~\ref{tab:seg}, it attains the best AUC on EX and the best F1 score on HE, indicating competitive preservation of fine-grained lesion structure.

\begin{table}[ht]
\centering
\resizebox{0.9\columnwidth}{!}
{
\begin{tabular}{c|cc|cc}
\hline
\textbf{Method}& \textbf{SGA} & \textbf{EVP} & \textbf{SSIM}$\uparrow$ & \textbf{PSNR}$\uparrow$ \\ \hline
\multirow{1}{*}{OTEGAN} & $\times$ & $\times$ & 0.9392 & 24.812 \\
\hline
\multirow{2}{*}{Ours (VAOT)}& $\checkmark$ & $\times$ & 0.9406 & 24.967 \\
&$\checkmark$ & $\checkmark$ & 0.9433 & 25.253 \\
\hline
\end{tabular}
}
\caption{Ablation Study of SGA and EVP structures. Specifically, OTEGAN denotes the backbone without any vessel-aware modules. Adding SGA and EVP progressively improves denoising (e.g., PSNR/SSIM), indicating their orthogonal effects. See Sec.~\ref{subsec:ablation} for details.}
\label{tab:ablation}
\end{table}
\subsection{Ablation Study}\label{subsec:ablation}
In VAOT, we introduce two structure-aware components to better preserve vascular anatomy during retinal fundus enhancement. Specifically, the SGA regularizer promotes global vessel-skeleton alignment, while the EVP regularizer encourages local consistency around detected endpoints. We assess their individual contributions with an ablation study on the same test set used for the main enhancement task. As shown in Tab.~\ref{tab:ablation}, adding SGA to the backbone (i.e., OTEGAN) yields modest gains in SSIM and PSNR. Adding EVP on top of SGA produces larger improvements, increasing SSIM from 0.9406 to 0.9433 and PSNR from 24.967 to 25.253, indicating that the two regularizers provide complementary benefits. Unless otherwise noted, all hyperparameters match those of the main experiments.

A natural question is why SGA alone offers only marginal improvements. We attribute this to a deliberately conservative weight $\lambda_s=1$ used across all experiments. Because exact skeleton alignment, can be brittle under noisy inputs, we position SGA as a gentle prior that biases the solution toward preserving global vessel morphology without enforcing exact, potentially unsafe matches. By contrast, the EVP regularizer delivers larger improvements because it directly targets local discontinuities around detected endpoints, no matter those endpoints arise from denoising gaps or imaging artifacts. As a result, it could strengthening local morphology and connectivity.

\section{Conclusion}
In this work, we present VAOT, a vessel-aware optimal transport framework for unpaired retinal fundus image enhancement. To preserve global vascular morphology, we introduce a skeleton guided morphology alignment regularizer. To maintain local vascular integrity, including connectivity and small loop consistency, we introduce a regularizer guided by detected endpoints and applied within local windows. Extensive experiments demonstrate that VAOT is effective and generalizes well for the primary enhancement task and across datasets. Moreover, VAOT better preserves vascular structure, as evidenced by improved performance in downstream vessel segmentation. We believe these findings advance structure preserving enhancement of unpaired fundus images and provide a useful foundation for future research.

{\small
\bibliographystyle{ieee_fullname}
\bibliography{egbib}
}

\end{document}


\title{Supplementary Materials - Bridging Restoration and Diagnosis: A Comprehensive Benchmark for Retinal Fundus Enhancement}

\author{First Author\\
Institution1\\
Institution1 address\\
{\tt\small firstauthor@i1.org}
\and
Second Author\\
Institution2\\
First line of institution2 address\\
{\tt\small secondauthor@i2.org}
}
\maketitle

\appendix
 \begin{figure*}[h]
  \centering
  \includegraphics[width=\textwidth]{ana_overall.png}  
  \caption{ Overview of EyeQ~\cite{fu2019evaluation} dataset. (A) highlights attribute distributions (i.e., brightness, contrast, sharpness) and diabetic retinopathy (DR) grades across quality categories (i.e., good, usable, and reject). (B) illustrates histograms for the training (i.e., part A and part B), testing, and validation datasets used in \textbf{Full-Reference} evaluations after resampling, with the workflow of degradation algorithms outlined below. (C) shows histograms for real-world \textbf{No-Reference} experiments after resampling. (D) presents reject-quality samples (e.g., overprocessed images).
  }
  \label{fig:overall_ana}
\end{figure*}

\section{Clinical Experts Guided Data Annotation Details}
We utilized 28,791 color fundus images from the EyePACS initiative~\cite{diabetic-retinopathy-detection}, with image quality labels obtained from the EyeQ dataset~\cite{fu2019evaluation}. Each image in the dataset~\cite{fu2019evaluation} was originally assigned a quality category (i.e., good, usable, or reject) and a diabetic retinopathy (DR) severity grade ranging from 0 to 4. As shown in Fig.~\ref{fig:overall_ana}(A), brightness, contrast, and sharpness distributions vary across quality levels, with good and usable images exhibiting similar patterns, while rejected images display distinct characteristics, such as increased sharpness. DR label distribution is imbalanced, with grades 0 and 2 being most frequent and severe cases underrepresented. Overprocessed images were also observed in usable and reject categories, potentially affecting diagnostic utility (Fig.~\ref{fig:overall_ana}(D)). To ensure quality and lesion preservation, we retained only good and usable images, applying ratio-preserving resampling under expert guidance. Due to the scarcity of severe DR cases, we preserved the natural DR label distribution to reflect clinical data characteristics.

\noindent \textbf{Full-Reference Evaluation Dataset.}
We used 16,817 good-quality images, split into 10,000 for training, 600 for validation, and 6,217 for testing (i.e., Fig.~\ref{fig:overall_ana}(B)). All images were synthetically degraded following~\cite{shen2020modeling}, simulating illumination issues, spot artifacts, and blurring. The training set was divided into two disjoint subsets (i.e., $A$ and $B$, each with 5,000 images), and corresponding degraded versions (i.e., $A^{\ast}$ and $B^{\ast}$) were generated. We strictly followed paired  (i.e., $A^{\ast} \rightarrow A$) and unpaired (i.e., $A^{\ast} \rightarrow B$) training pipeline for fair comparison.

\noindent \textbf{No-Reference Evaluation Dataset.}
We selected 6,434 usable-quality images (i.e., Fig.~\ref{fig:overall_ana}(C)), resampling 4,000 for training and 2,434 for testing under real-world noise conditions. Additionally, 4,000 unpaired good-quality images were resampled from the original training pool (i.e., used in Full-Reference Evaluation), with DR label matching to support unpaired training protocols.

\section{Full-Reference Quality Assessment Experiments Details }\label{Sec:full-reference}

For full-reference assessment, we used the previously synthesized Full-Reference Evaluation Dataset. We strictly followed the training configurations for paired and unpaired methods. For the unpaired method, synthetic low-quality images from the training set $A$ (i.e., $A^{\ast}$) were used as input images, while high-quality images from the training set $B$ served as the clean reference images. For the paired method, we performed supervised training using low-high-quality image pairs from the training set $A$ (i.e., $A^{\ast}$ and $A$).

\subsection{SCR-Net~\cite{li2022structure}}
The model was trained for 150 epochs using Adam optimizer, with an initial learning rate of $2 \times 10^{-4}$ and $\beta_1$ value set to $0.5$, followed by 50 epochs with a learning rate linearly decayed to $0$. The training batch size was 32. All images were resized to $ 256 \times 256$ with a random flipping data augmentation technique. For model architectures, the generator and discriminator architectures followed the architectures and configurations described in~\cite{li2022structure}.

\subsection{Cofe-Net~\cite{shen2020modeling}}

The model was trained for 300 epochs using the SGD optimizer, with an initial learning rate of $1 \times 10^{-4}$, which was gradually reduced to 0 over the final 150 epochs. The training batch size was 16, and all images were resize to $512 \times 512$.

The loss function comprised four components: main scale error loss ($L_m$), multiple-scale pixel loss ($L^s_p$), multiple-scale content loss ($L^s_c$) and RSA module loss ($L_v$), as described in~\cite{shen2020modeling}, where the $s$ denotes the scale index. The weight for $L^s_p$, $L^s_c$ and $L_v$ was set to $\lambda_p=10$, $\lambda_c=1$ and $\lambda_v=0.1$, respectively, during the training process.

\subsection{PCE-Net~\cite{10.1007/978-3-031-16434-7_49}}

The model was trained for 200 epochs using the Adam optimizer, with an initial learning rate of $1 \times 10^{-3}$, which was gradually reduced to 0 over the final 50 epochs. The training batch size was 4, and all input images were resized to $256 \times 256$. Data augmentation strategies, including random horizontal and vertical flips with a probability of 0.5, were applied to enhance generalization.

The loss function comprised two components: enhancement loss ($L_E$) and the weighted feature pyramid constraint loss ($L_C$), as described in~\cite{10.1007/978-3-031-16434-7_49}. The weight for $L_C$ was set to $\lambda_C=0.1$ during the training process. Additionally, we adopted a U-Net architecture proposed in ~\cite{10.1007/978-3-031-16434-7_49}.

\subsection{GFE-Net~\cite{li2023generic}}

The model was trained for 200 epochs using the Adam optimizer, with an initial learning rate of $1\times 10^{-3}$, which was gradually reduced to 0 over the final 50 epochs. The training batch size was set to 4, and all input images were resized to $256 \times 256$. Data augmentation strategies, including random horizontal and vertical flips with a probability of 0.5, were applied to enhance generalization.

We employed the same weight (e.g., $\lambda_{all}$ = 1) for all loss losses, including enhancement loss, cycle-consistency loss, and reconstruction loss. Furthermore, we adopted the architecture proposed in~\cite{li2023generic}, implementing a symmetric U-Net with 8 down-sampling and 8 up-sampling layers.

\subsection{I-SECRET~\cite{i-secret}}

The model was trained for 200 epochs using Adam optimizer with an initial learning rate of $1 \times 10^{-4}$ and $\beta$ values set to $0.5$ and $0.999$, respectively. The learning rate followed a cosine decay schedule. The training batch size was set to 8. All images were resized to $256 \times 256$ with random cropping and flipping augmentation strategies.

For model architectures, the generator consisted of 2 down-sampling layers, each with 64 filters and 9 residual blocks. Input and output channels were set to 3 for RGB inputs. The discriminator included 64 filters and 3 layers. Instance normalization and reflective padding were used. The training process employed a least-squares GAN loss~\cite{mao2017least}, a ResNet-based generator, and a PatchGAN-based~\cite{isola2017image} discriminator. GAN and reconstruction losses were weighted at $1.0$, while their importance with the contrastive loss (ICC-loss) and importance-guided supervised loss (IS-loss)~\cite{i-secret} were enabled with weights of $1.0$.

\subsection{ RFormer~\cite{deng2022rformer}.}
The model was trained for 150 epochs using Adam optimizer, with an initial learning rate of $1 \times 10^{-4}$ and $\beta$ values set to 0.9 and 0.999, respectively. The cosine annealing strategy was employed to steadily decrease the learning rate from the initial value to $1 \times 10^{-6}$ during the training procedure. The training batch size was set to 32. All images were resized to $256 \times 256$ without any additional augmentation strategies. The model architecture followed the design proposed in~\cite{deng2022rformer}, which was consistently maintained throughout our experiments.

\subsection{CycleGAN~\cite{cyclegan}, WGAN~\cite{gulrajani2017improved}, OTTGAN~\cite{wang2022optimal}, OTEGAN~\cite{zhu2023optimal}  }
The models were trained for 200 epochs using the RMSprop optimizer, with initial learning rates for the generator and discriminator set to $0.5 \times 10^{-4}$ and $1 \times 10^{-4}$, respectively. The learning rate followed a linear decay schedule, decreasing by a factor of 10 every 100 epochs. The training batch size was set to 2. All input images were resized to $256 \times 256$, with random horizontal and vertical flips applied as augmentation strategies.
For CycleGAN, the weighting parameters in the final objective were set to $\lambda_{GAN} = 1$, $\lambda_{Cycle} = 10$, and $\lambda_{Idt} = 5$, corresponding to the weights for the GAN loss, cycle consistency loss, and identity loss, respectively. The Mean Squared Error (MSE) loss was used for the GAN loss, while the cycle consistency and identity losses were computed using the L1-norm. For OTTGAN and OTEGAN, the weighting parameter $\lambda_{OT}$ was set to 40, representing the optimal transport (OT) cost. Furthermore, the OT loss was calculated using the MSE loss for OTTGAN and the MS-SSIM loss for OTEGAN. The generator and discriminator architectures were implemented following the baseline designs described in~\cite{zhu2023optimal,zhu2023otre}.

\subsection{Context-aware OT~\cite{vasa2024context}}
The model was trained for 200 epochs using the RMSprop optimizer, with initial learning rates for the generator and discriminator set to $0.5 \times 10^{-4}$ and $1 \times 10^{-4}$, respectively. The learning rate followed a linear decay schedule, decreasing by a factor of 10 every 50 epochs. The training batch size was set to 2. All input images were resized to $256 \times 256$ without additional augmentation strategies.
A warm-up training strategy was employed, wherein the context-OT loss was introduced after the first 50 epochs. The weighting parameter for this loss was set to $5\times10^{-2}$. We utilized a pre-trained VGG~\cite{mechrez2018contextual} network outlined in~\cite{vasa2024context} to compute the OT loss at feature spaces.
The generator and discriminator architectures followed the designs outlined in~\cite{vasa2024context}.

\subsection{TPOT~\cite{dong2024tpot}}
The model was trained for 100 epochs using the RMSprop optimizer over two training phases. In each phase, the learning rate was initialized at $2 \times 10^{-4}$ and reduced by a factor of 10 after
every $50$ epochs. The training batch size was set to 4, and all input images were resized to $256 \times 256$ without additional augmentation strategies. The weighting parameter for the topology regularization was fixed at 1. where the segmentation masks During training, the segmentation masks were extracted using the method proposed in~\cite{zhou2021study}. The generator and discriminator architectures followed the designs introduced in~\cite{dong2024tpot}.

\subsection{CUNSB-RFIE~\cite{dong2024cunsb}}
The model was trained for 130 epochs using the Adam optimizer, with an initial learning rate of $2 \times 10^ {-4}$. The learning rate was linearly decayed to 0 after the first 80 epochs, and the batch size was set to 8. All input images were resized to $256 \times 256$ without applying any additional augmentation strategies.

The weighting parameters in the final objective were set as $\lambda_{SB} = 1$, $\lambda_{SSIM} = 0.8$, and $\lambda_{NCE} = 1$, corresponding to the weights for entropy-regularized OT loss, task-specific regularization with MS-SSIM~\cite{brunet2011mathematical}, and PatchNCE~\cite{park2020contrastive} loss, respectively.

The generator and discriminator architectures followed the designs described in~\cite{dong2024cunsb}. Specifically, the base number of channels for the generator was set to 32, and 9 ResNet blocks were used in the bottleneck. In addition to the output features of all downsampling layers, the bottleneck's input and middle feature maps were utilized to calculate the PatchNCE regularization.

\subsection{Vessel Segmentation}
A vanilla U-Net model~\cite{ronneberger2015unet} was employed for the downstream vessel segmentation task. The network comprised 4 layers with a base channel size 64 and a channel scale expansion ratio of 2. The training was conducted over 10 epochs using the Adam optimizer, with a batch size of 64 and an initial learning rate of $5 \times 10^{-5}$, which followed a cosine annealing learning rate scheduler. 

Before training, the enhanced images and their corresponding ground-truth vessel segmentation masks were preprocessed. The preprocessing pipeline included random cropping to $ 48 \times 48$ patches, followed by data augmentation techniques such as random horizontal flips, random vertical flips (with a probability of 0.5), and random rotation.

\begin{figure}[ht]
    \centering
    \includegraphics[width=1\linewidth]{experimental_design.png}
    \caption{An illustrative medical expert clinical preference evaluation between (a) lesion preserving, (b) background preserving, and (c) structure-preserving.}
    \label{fig:expert-protocol}
\end{figure}

\begin{figure*}[t]
  \centering
  \includegraphics[width=\textwidth]{fig_eyeq.pdf}  
  \caption{ Illustration of the Denoising Evaluation on the EyeQ dataset. The first and second columns show the high- and low-quality image references, respectively, while the remaining columns display the synthetic high-quality images generated by all baseline models.
  }
  \label{fig:full-reference-eyeq}
\end{figure*}
\begin{figure*}[t]
  \centering
  \includegraphics[width=\textwidth]{fig_other.pdf}  
  \caption{ Illustration of the Denoising Generalization Evaluation on the DRIVE and IDRID datasets. The first and second columns show the high- and low-quality image references, respectively, while the remaining columns display the synthetic high-quality images generated by all baseline models.
  }
  \label{fig:full-reference-generalization}
\end{figure*}
\begin{figure*}[t]
  \centering
  \includegraphics[width=\textwidth]{fig_seg.pdf}  
  \caption{ Illustration of Vessel and Lesion (EX and HE) Segmentation Experiments. The first column shows the reference segmentation masks, while the remaining columns display the segmentation results produced by all baseline models.
  }
  \label{fig:full-reference-segmentation}
\end{figure*}
\begin{figure*}[t]
  \centering
  \includegraphics[width=\textwidth]{fig_noref.pdf}  
  \caption{ Illustration of the denoising results in the No-Reference Quality Assessment Experiments. The first column shows the input low-quality image, while the remaining columns display the synthetic high-quality images generated by all unpaired baseline models.
  }
  \label{fig:no-reference}
\end{figure*}

\section{No-Reference Quality Assessment Experiments Details}
We utilized the No-Reference Evaluation Dataset, including all unpaired baseline models for the No-Reference Assessment. These experiments evaluated the models' ability to learn and eliminate real-world noise. We maintained the experimental settings (e.g., hyperparameters) as outlined in Sec.~\ref{Sec:full-reference} to ensure a fair comparison.

\subsection{Lesion Segmentation}
Another U-Net model was employed for the downstream lesion segmentation task. The network consisted of 4 layers, with a base channel size of 64. The channel multiplier was set to 1 in the final layer and 2 in the remaining layers. The model was trained for 300 epochs using the Adam optimizer, with a batch size of 8. The initial learning rate was set to $2 \times 10^{-4}$, and a cosine annealing scheduler was applied, gradually reducing the learning rate to a minimum value of $1 \times 10^{-6}$. 

We utilized extensive data augmentation strategies to enhance model robustness. These included random horizontal and vertical flips, each with a probability of 0.5; random rotations with a probability of 0.8; random grid shuffling over $8\times 8$ grids with a probability of 0.5; and CoarseDropout, which masked up to 12 patches of size 
$20 \times 20$ to a value of 0, also with a probability of 0.5.

\subsection{DR grading.} We trained an NN-MobileNet model~\cite{deeplearning1} for the DR grading task using real-world high-quality images. The enhanced test images are used with the trained NN-MobileNet to infer DR grading classification. Enhancement performance is evaluated based on classification accuracy (ACC), kappa score, F1 score, and AUC. This evaluation primarily aims to assess whether the denoising model disrupts lesion distribution, potentially leading to inconsistencies with the original DR grading labels. 
During the training, we conducted 200 epochs with a batch size of 32 and an input size of $256 \times 256$. The AdamP optimizer was utilized with a $1 \times 10^{-3}$ weight decay and an initial learning rate of $1 \times 10^{-3}$. A dropout rate of $0.2$ was applied during training to mitigate over-fitting. Furthermore, the learning rate was dynamically adjusted using the Cosine Learning Rate Scheduler.

\subsection{Representation Feature Evaluation.}We employed two foundation models for fundus images, RetFound~\cite{zhou2023foundation} and Ret-CLIP~\cite{du2024ret}, to calculate the Fréchet Inception Distance (FID) between enhanced and real-world high-quality image feature representations. These metrics are referred to as \textit{FID-RetFound} and \textit{FID-CLIP}, respectively.

\textit{FID-Retfound}, based on a MAE backbone, captures high-level semantic structures, while \textit{FID-Clip}, trained via contrastive learning, emphasizes spatial coherence and structural consistency. To compute these metrics, the enhanced and real-world high-quality images were resized to $224 \times 224$ and normalized before being passed into the respective image encoders. The FID scores were then calculated based on the 1024-dimensional and 512-dimensional feature maps produced by RetFound and Ret-Clip, respectively.

\subsection{Experts Annotation Evaluation.} To evaluate the quality of the enhanced images, we recruited six trained specialists to conduct manual assessments. The evaluation criteria, as illustrated in Fig.~\ref{fig:expert-protocol}, included lesion preservation, background preservation, and structure preservation. Each image was individually reviewed, and the results were meticulously recorded. To minimize subjective bias, the six annotators performed cross-evaluations on test images enhanced by different models. The final annotations were further validated by ophthalmologists to ensure accuracy and clinical relevance. Despite these efforts, a degree of variability may still persist due to the inherent subjectivity and manual nature of expert evaluations.

\section{Result Illustrations}
We provide additional visualizations for all baseline models in the Full-Reference and No-Reference Quality Assessment Experiments. Specifically, Fig.~\ref{fig:full-reference-eyeq} presents the results of the Denoising Evaluation conducted on the EyeQ dataset. In contrast, Fig.~\ref{fig:full-reference-generalization} illustrates the Denoising Generalization Evaluation results on the DRIVE~\cite{drive} and IDRID~\cite{idrid} datasets. Fig.~\ref{fig:full-reference-segmentation} displays the outcomes of Vessel and Lesion (EX and HE) Segmentation. The results of the No-Reference quality assessment Experiments are outlined in Fig.~\ref{fig:no-reference}.

{\small
\bibliographystyle{ieee_fullname}
\bibliography{egbib}
}